\documentclass[conference]{IEEEtran}
\IEEEoverridecommandlockouts

\usepackage{cite}
\usepackage{amsmath,amssymb,amsfonts}
\usepackage{algorithmic}
\usepackage{graphicx}
\usepackage{textcomp}
\usepackage{xcolor}
\usepackage{booktabs}
\usepackage{multirow}
\usepackage{array}
\usepackage{url}
\usepackage{microtype}
\usepackage{hyperref}
\usepackage{balance}

\def\BibTeX{{\rm B\kern-.05em{\sc i\kern-.025em b}\kern-.08em
    T\kern-.1667em\lower.7ex\hbox{E}\kern-.125emX}}

\begin{document}

\title{
Benchmarking Google Embeddings 2 against Open-Source Models for Multilingual Dense Retrieval and RAG Systems
}

\author{
\IEEEauthorblockN{Stefano Cirillo}
\IEEEauthorblockA{\textit{University of Salerno} \\
Fisciano, Salerno, Italy}
\and

\IEEEauthorblockN{Domenico Desiato}
\IEEEauthorblockA{\textit{University of Bari} \\
Bari, Italy}
\and

\IEEEauthorblockN{Giuseppe Polese}
\IEEEauthorblockA{\textit{University of Salerno} \\
Fisciano, Salerno, Italy}
\and

\IEEEauthorblockN{Giandomenico Solimando}
\IEEEauthorblockA{\textit{University of Salerno} \\
Fisciano, Salerno, Italy}
}

\maketitle

\begin{abstract}
We benchmark Google Embeddings  (GE2), a Vertex-AI-hosted
bi-encoder with 2{,}048-token context and explicit task-type
conditioning, against five open-source alternatives: BGE-M3, E5-large,
Multilingual-E5-large (mE5-L), LaBSE, and Paraphrase-Multilingual-MPNet
(mMPNet). Evaluation covers four BEIR subsets, a synthetic Italian RAG
corpus, a chunking
ablation considering 5 sizes of tokens with three strategies,
and per-query latency on commodity CPU hardware. GE2 ranks first on
every task, achieving BEIR avg.\ nDCG@10 = 0.638 and IT-RAG-Bench
nDCG@10 = 0.282, but at 231.6~ms median latency it is roughly
$14\times$ slower than the fastest local models. mE5-L reaches
within 0.003 nDCG of GE2 on Italian at 31~ms, making it the
preferred option when sub-100~ms SLAs matter. A more striking finding
concerns LaBSE, which despite widespread multilingual deployment
scores 0.188 average nDCG@10 on BEIR, below every dedicated retrieval
model including mMPNet. Chunking experiments show that all six models
saturate at 32-token chunks on our corpus, with semantic chunking
providing measurable gains only at 16 tokens. Code and datasets are
publicly released\footnote{The code is publicly available at: \href{https://github.com/cciro94/GoogleEmbeddings2-benchmark}{https://github.com/cciro94/GoogleEmbeddings2-benchmark}}.
\end{abstract}


\section{Introduction}

Retrieval-augmented generation (RAG)~\cite{lewis2020rag} has emerged as the predominant paradigm for grounding large language models in external knowledge sources. However, the performance of any RAG system is fundamentally constrained, in a rigorous information-theoretic sense, by the effectiveness of its retrieval component. Any passage that is not retrieved is unavailable for citation, summarization, or downstream reasoning, independent of the capabilities or expressiveness of the generative model~\cite{shi2023large}. This bottleneck property has an important practical implication that is often underestimated: errors occurring at the retrieval stage are irrecoverable for all subsequent processing, whereas errors introduced during generation remain observable and can, in principle, be identified and corrected by a human reader. Ensuring correct retrieval is therefore not merely a matter of performance optimization, but constitutes a necessary precondition for the downstream system to function at all.

The predominant architecture for scalable dense retrieval is the bi-encoder~\cite{karpukhin2020dense}, which independently maps queries and documents into a shared vector space and estimates their relevance via cosine similarity. This design enables offline indexing of the document corpus, after which subsequent queries can be resolved in near-constant time using approximate nearest neighbor (ANN) search~\cite{johnson2019faiss}, allowing the method to scale to hundreds of millions of passages in production environments, characteristics that are well documented in the literature. What remains comparatively underexplored, and is the primary focus of this paper, is the manner in which the selection of an embedding model interacts with the operational constraints of a deployed RAG pipeline. These constraints include language coverage, maximum document length, chunking granularity, latency requirements, and the inherent structural mismatch between short user queries and substantially longer retrieved passages.

Three aspects of this problem are particularly underappreciated in
practitioner workflows. The first is the distinction between
\emph{symmetric} and \emph{asymmetric} similarity objectives.
Many widely used multilingual models, including LaBSE~\cite{feng2022language},
were trained to align parallel sentences across languages, a task that
requires a symmetric distance measure and does not involve any notion
of topical relevance between a short query and a longer passage.
Applying such models to retrieval treats a fundamentally classification-
or alignment-like objective as a ranking objective, and the two are not
interchangeable. Despite this mismatch, LaBSE has accumulated over 10M
downloads on HuggingFace and is routinely cited in documentation and
tutorials as a general-purpose multilingual retrieval model. Our results
suggest this usage pattern is a systematic error with measurable quality
consequences.

A second, often underrecognized, consideration is context length. The majority of publicly available open-source embedding models impose a maximum input size of 512 subword tokens, a limitation inherited from the BERT-family architectures on which they are based. In practical RAG settings, however, many corpora—especially in legal, medical, and administrative domains—contain passages that substantially exceed this bound. The prevailing mitigation strategy is truncation, which silently discards semantically pertinent content and thereby degrades retrieval performance in proportion to the fraction of relevant information lying beyond the cutoff. Google’s Embeddings (GE2) increases this context budget to 2,048 tokens, but, to our knowledge, the resulting effect on retrieval quality has not yet been independently evaluated or systematically characterized.

The third aspect is chunking. When documents exceed the model's context
window, they must be split into smaller units before embedding.
The literature contains advice ranging from fixed-size splits at
256 or 512 tokens to sophisticated semantic segmentation, but
systematic evidence for which strategy matters and at what granularity
remains sparse. Practitioners typically tune chunking by intuition or
from general guidelines rather than from dataset-specific ablations,
and the interaction between chunk size and model architecture is rarely
examined.

Against this background, we present a systematic benchmarking study
of GE2 alongside five open-source alternatives. Our central goal is
not to crown a winner, which is straightforward once the experiments
are run, but to understand \emph{why} models diverge in quality, where
their failure modes lie, and what deployment constraints determine
whether the quality gap justifies the latency cost. We find that GE2
achieves consistently better retrieval quality, especially on tasks
involving long documents or strongly heterogeneous query-document
surface forms, but that mE5-L closes almost the entire gap on
short-passage, single-language corpora at a fraction of the latency.
The findings directly inform model selection decisions that practitioners
must make under real constraints.

Our main contributions are summarized as follows:
\begin{enumerate}
    \item We present a fully reproducible evaluation of GE2 against five open-source models on BEIR, IT-RAG-Bench (a newly introduced synthetic Italian corpus), and under a systematic sweep of chunking strategies.
    \item We provide, to the best of our knowledge, one of the first benchmarks comparing GE2 with multilingual open models for Italian passage retrieval.
    \item We offer an empirical characterization of chunking sensitivity across all six models, revealing a saturation plateau at 32 tokens for short-passage corpora and a pronounced performance degradation below 16 tokens.
    \item We conduct a latency–quality Pareto analysis that yields actionable guidance for model selection in RAG deployments under heterogeneous SLA constraints.
\end{enumerate}

\section{Related Work}

\subsection{Retrieval Benchmarks and Evaluation Protocols}

Standardized benchmarks have been essential for separating genuine
improvements in dense retrieval from in-distribution overfitting.
BEIR~\cite{thakur2021beir} established the current standard by
aggregating 18 heterogeneous retrieval tasks spanning biomedical,
financial, legal, and scientific domains under a zero-shot evaluation
protocol, where models are not fine-tuned on target datasets. The
zero-shot condition is critical: it exposes models whose strong
in-domain training prevents generalization, a failure mode that affects
several high-capacity models more than might be expected from their
within-domain scores.

MTEB~\cite{muennighoff2022mteb} further broadens the evaluation landscape to 56 tasks, encompassing clustering, classification, and semantic textual similarity in addition to retrieval. Although MTEB provides a broad assessment of embedding utility, aggregating results across diverse task types can hide a model’s specific retrieval behavior. Models that achieve state-of-the-art performance on classification tasks may exhibit suboptimal behavior on ranking tasks, and conversely, models optimized for ranking performance may underperform on classification.

Multilingual retrieval is addressed by MIRACL~\cite{zhang2023miracl},
which covers 18 languages and was specifically designed to avoid
reliance on English-centric evaluation assumptions. A consistent finding
from MIRACL is that performance drops substantially on non-English
tasks even for models with explicit multilingual training, and that
this degradation is uneven across models and language families.
Our IT-RAG-Bench is designed in a similar spirit but is specifically adapted to the retrieval-augmented generation (RAG) setting, featuring passage lengths and domain distributions (Wikipedia, public administration, civil code) that differ substantially from MIRACL’s news-oriented corpora.

\subsection{Bi-Encoder Architectures and Training Objectives}

The architecture underlying most modern dense retrievers is a dual
encoder, or bi-encoder, in which separate (or weight-shared) encoders
produce query and document representations, scored by dot product or
cosine similarity~\cite{karpukhin2020dense}. Within this broad
category, training objectives vary considerably and have significant
consequences for downstream behavior.

\textbf{Contrastive retrieval training.}
DPR~\cite{karpukhin2020dense} introduced the now-standard recipe of
training on (query, positive passage, hard negative) triples with
in-batch negatives. Its key insight was that hard negatives, mined
from BM25 or previous model versions, are necessary to prevent the
model from learning a trivially separable embedding space that does
not capture fine-grained relevance. Subsequent work has refined this
recipe substantially. E5~\cite{wang2022text} augments contrastive
training with weak supervision from (title, body) pairs scraped from
the web, yielding strong zero-shot generalization without requiring
labeled retrieval datasets. This approach reduces the dependency on
expensive annotation pipelines and scales naturally to new languages
when multilingual web data is available, as in mE5-L.

\textbf{Multi-task and multi-granularity models.}
BGE-M3~\cite{chen2024bge} trains a single model to support dense
retrieval, sparse retrieval, and multi-vector (ColBERT-style)
retrieval simultaneously, using self-knowledge distillation to align
the three output heads. The motivation is compelling in principle:
a unified model could adapt its retrieval mode depending on the
downstream application. In practice, however, multi-task training
involves optimization tradeoffs that do not uniformly benefit all
tasks. Our results show that BGE-M3 underperforms mE5-L by a
substantial margin on BEIR zero-shot retrieval despite its larger
parameter count, a pattern consistent with the multi-task tension
between contrastive ranking and the other objectives.

\textbf{Sentence similarity models repurposed for retrieval.}
LaBSE~\cite{feng2022language} was designed for language-agnostic
sentence embedding, trained primarily on bitext pairs for cross-lingual
sentence alignment. Its training signal is symmetric: both elements
of a pair are ``similar'' sentences in different languages, and the
model learns a distance function that is invariant to surface form
differences. This is precisely the wrong inductive bias for passage
retrieval, where the query is typically a short natural-language
question and the relevant passage is a longer document segment with
different vocabulary, style, and structure. The model has no training
signal to align these heterogeneous objects, and its embeddings
reflect sentence-level proximity rather than topical relevance.
The observed mismatch does not constitute a deficiency in LaBSE relative to its originally intended applications. Rather, employing LaBSE as the primary retrieval backbone in RAG pipelines represents a fundamental methodological misalignment, one that is, nonetheless, frequently encountered in practitioner-oriented tutorials and publicly available code repositories.

A similar issue affects Sentence-BERT variants trained with paraphrase-oriented objectives, such as mMPNet. These models are most commonly assessed on semantic textual similarity (STS) benchmarks, in
which near-duplicate detection constitutes the predominant evaluation scenario. However, STS scores exhibit weak correlation with BEIR retrieval performance because the underlying distributions of
positive pairs differ substantially: paraphrase pairs typically share substantial surface-level
lexical overlap with their queries, whereas in information retrieval, relevant passages often do not.
Consequently, paraphrase-optimized models fail to learn the asymmetric relevance patterns needed for effective retrieval.

\textbf{API-hosted models with asymmetric task conditioning.}
GE2 (Embeddings) departs from the conventional symmetric bi-encoder paradigm by applying distinct representation transformations conditioned on whether the input is designated as a query
(\texttt{RETRIEVAL\_QUERY}) or as a document (\texttt{RETRIEVAL\_DOCUMENT}). This asymmetric conditioning is realized at the task-type level, rather than via separate encoder parameterizations, yet it effectively induces a systematic displacement of query and document representations toward geometrically complementary regions of the embedding space. The \texttt{text-embedding-3} series from OpenAI adopts a conceptually similar strategy, and preliminary internal assessments from both providers indicate that asymmetric conditioning produces consistent performance improvements on tasks in which queries and documents exhibit divergent surface-level statistical properties. To date, no independent zero-shot evaluation of GE2 relative to open-source alternatives has been reported prior to the present study.

\subsection{Chunking Strategies in Dense Retrieval}

Chunking, i.e., the process of partitioning long documents into smaller, independently retrievable units, has received comparatively limited attention in the research literature relative to model architecture and training methodologies, despite constituting a practical deployment requirement for nearly all real-world RAG systems. The central design trade-off concerns information density versus boundary-induced fragmentation effects. Larger chunks better preserve coherence across sentences, but can reduce retrieval precision by including extraneous or weakly relevant context. Smaller chunks focus task-relevant information and can improve retrieval specificity, but they increase the risk that an argument, narrative segment, or factual claim is split across chunk boundaries, potentially impairing later reasoning and generation.

Fixed-size partitioning with optional overlap represents the most elementary segmentation strategy and is commonly used as a baseline in comparative evaluations. Sliding-window variants introduce controlled overlap between consecutive segments to decrease the likelihood that a semantically relevant span is divided across boundaries, albeit at the cost of approximately doubling or tripling the index size and increasing redundancy in the retrieved results. TextTiling-inspired semantic Segmentation~\cite{hearst1997texttiling} seeks to detect natural topical boundaries by identifying drops in cosine similarity between adjacent sentence embeddings and subsequently aggregating sentences into thematically coherent segments. This method is more aligned with discourse structure but incurs a dependency on the quality of sentence-level embeddings and exhibits sensitivity to the choice of similarity-threshold hyperparameter.

\cite{rau2024context} show that different embedding models degrade
non-uniformly as chunk length increases beyond the model's effective
context budget, motivating model-specific chunking decisions.
We extend this line of inquiry in two directions: toward very short
chunks, where a different failure mode, semantic fragmentation below
coherence threshold, becomes the dominant risk, and to a multilingual
short-passage corpus where we identify an early and universal
saturation effect that has not been previously reported.

\section{Methodology}

\subsection{Overview of the Evaluation Pipeline}

Our evaluation follows a standard offline retrieval protocol, but
spans several distinct stages, each of which introduces design choices
that we describe explicitly here to ensure reproducibility and to
clarify the source of any performance differences. At a high level,
the pipeline proceeds as follows: documents from each corpus are
chunked according to the strategy and size under evaluation, each
chunk is embedded using the target model, the resulting vectors are
indexed via FAISS, each query is embedded and used to retrieve the top
$k$ chunks, and retrieved chunks are mapped back to their parent
documents for evaluation against ground-truth relevance labels.

The mapping from chunk-level retrieval to document-level relevance
is a non-trivial design decision. We assign a chunk the relevance
label of its parent document, which means that a correctly retrieved
chunk from a relevant document counts as a hit regardless of which
chunk within that document it came from. This is appropriate for our
IT-RAG-Bench corpus, where passages are short enough that most
chunks are near-complete passages, but it introduces a mild
optimistic bias for fixed chunking at large sizes, where a single
relevant passage may generate fewer chunks than a non-relevant one.
We discuss the magnitude of this effect in the chunking ablation
(Section~\ref{sec:chunkresults}).

\subsection{Retrieval Scoring}

A bi-encoder $\phi:\mathcal{T}\!\rightarrow\!\mathbb{R}^k$ ranks
documents in corpus $\mathcal{D}=\{d_i\}_{i=1}^N$ by
\begin{equation}
  \text{sim}(q,d_i)=\frac{\phi(q)^\top\phi(d_i)}
    {\|\phi(q)\|_2\cdot\|\phi(d_i)\|_2}\,.
  \label{eq:cosine}
\end{equation}
All corpus representations are computed once in a preprocessing phase prior to any evaluation.  
For GE2, document-side embeddings are produced using the task specification \texttt{RETRIEVAL\_DOCUMENT}, whereas query-side embeddings are produced using \texttt{RETRIEVAL\_QUERY}. All open-source models employ a single shared encoder for both queries and documents; for the E5 model variants, we prepend instruction prefixes (``query:'' / ``passage:'') in accordance with the procedures described in their original training protocols.

Pre-computed embeddings are persisted on disk and indexed by the key
SHA-256(\textit{model}\,$\|$\,\textit{text}\,$\|$\,\textit{normalize}),
which enables recovery and continuation of corpus indexing without reissuing API requests following partial system failures. This design is particularly critical for GE2, where API calls are rate-limited and repeated calls incur high computational and monetary costs.

Index construction is implemented using FAISS~\cite{johnson2019faiss} with an HNSW configuration
($M=32$, $\mathit{ef}_{\mathrm{c}}=200$, $\mathit{ef}_{\mathrm{s}}=100$) for collections with $N\geq10^5$ passages. In the present work, however, all corpora satisfy $N\leq5{,}000$, and we therefore employ exhaustive flat search in all experiments. This guarantees that the retrieval ranking is not influenced by approximation artifacts introduced by ANN methods. The choice is driven by corpus scale: while the trade-off between exact and approximate search is critical for large-scale collections, it is negligible for the dataset sizes considered here.

\subsection{Evaluation Metrics}
\label{sec:metrics}

We report three metrics that capture complementary aspects of
retrieval quality. Recall@$k$ measures the fraction of relevant
passages recovered in the top-$k$ retrieved results:
\begin{equation}
  \operatorname{Recall}@k=
  \frac{|\mathcal{R}_q\cap\hat{\mathcal{D}}_k|}{|\mathcal{R}_q|}\,,
  \label{eq:recall}
\end{equation}
where $\mathcal{R}_q$ is the set of relevant passages for query $q$
and $\hat{\mathcal{D}}_k$ is the top-$k$ retrieved set. Recall is
the most directly actionable metric for RAG, since it measures how
much of the retrievable answer pool is actually surfaced to the
generator.

Mean Reciprocal Rank measures top-1 precision by averaging the
reciprocal of the first relevant passage's rank:
\begin{equation}
  \operatorname{MRR}=\frac{1}{|Q|}\sum_{q\in Q}
  \frac{1}{\operatorname{rank}_q^{\mathrm{first}}}\,.
  \label{eq:mrr}
\end{equation}
MRR is informative because in single-answer retrieval tasks, placing
the correct passage at rank 1 versus rank 5 has a large impact on
generator quality even when both configurations yield the same Recall@10.

Our primary metric is nDCG@10, which applies a log-discount to
graded relevance scores $r_i\in\{0,1,2\}$, penalizing models that
place relevant passages at lower ranks within the top 10:
\begin{equation}
  \operatorname{nDCG}@k=\frac{\operatorname{DCG}@k}
    {\operatorname{IDCG}@k},\quad
  \operatorname{DCG}@k=\sum_{i=1}^{k}\frac{2^{r_i}-1}{\log_2(i+1)}\,.
  \label{eq:ndcg}
\end{equation}
nDCG@10 is the standard metric on BEIR and allows direct comparison
with published baselines. MRR and Recall@$k$ provide complementary
views on top-1 precision and coverage, respectively, and are
particularly useful for interpreting performance differences on
IT-RAG-Bench, where graded relevance is not available at scale.

\subsection{Chunking Strategies}
\label{sec:chunking}

We evaluate three strategies at five target sizes
$L\in\{8,16,32,64,128\}$ whitespace tokens.

\textit{Fixed} splits each document into non-overlapping contiguous
segments of exactly $L$ tokens, discarding a trailing fragment if
it falls below $L/4$ tokens. This is the simplest strategy and
serves as the baseline. Its failure mode is deterministic: clause
and sentence boundaries are ignored, so any coherent unit that
crosses a chunk boundary is fragmented.

\textit{Sliding window} uses the same fixed segmentation but with
a stride of $L/2$, so that adjacent chunks overlap by 50\%. This
doubles the index size relative to fixed chunking and introduces
redundancy in retrieval results (two chunks from the same region of
a document may both be retrieved), but it reduces the probability
that a relevant span is bisected at a hard boundary. We include it
primarily to provide a ceiling on how much of the fixed-chunking
deficit is recoverable by simple overlap without any semantic
knowledge.

\textit{Semantic} chunking detects boundaries where cosine similarity
between adjacent sentence embeddings falls below a threshold
$\tau=0.75$, following the TextTiling~\cite{hearst1997texttiling}
paradigm. Segment length is bounded to $[L/2,\,2L]$ tokens; if a
detected boundary would produce a segment outside this range, the
strategy falls back to a fixed split. Sentence embeddings for
boundary detection are computed using the same model being evaluated,
so semantic chunking is implicitly model-specific: the same text may
be segmented differently depending on which model's similarity
function determines the boundaries. This introduces a dependency
that is worth noting when interpreting results at smaller chunk sizes,
where the segmentation and the retrieval signal are more tightly
coupled.

Chunk-level relevance annotations are inherited from their corresponding source documents, as previously described. All retrieval strategies and chunk sizes are evaluated on the identical IT-RAG-Bench corpus using the same query set and relevance labels, thereby enabling direct and comparable nDCG-based performance assessments across experimental conditions.

\section{Experimental Setup}

\subsection{Models}

The six models are summarized in Table~\ref{tab:models}. GE2 is the
sole API-based model, accessed via the Vertex AI SDK
(\texttt{google-genai}) with Application Default Credentials; all
others run locally on Apple M-series CPU. GE2's two most architecturally
significant properties are its 2{,}048-token context window, which is
four times longer than all local models, and its asymmetric task-type
conditioning. The former matters most when passages are long relative
to the 512-token limit of BERT-based models; the latter matters most
when query and document surface statistics diverge substantially, as
in domain-heterogeneous corpora.

Among the open-source models, E5-large and mE5-L share the same
training recipe but differ in language coverage: E5-large is
English-only, while mE5-L was trained on multilingual data using
the same contrastive objective. Both use the instruction prefix
convention (``query:'' and ``passage:'') during inference, which is
a lightweight analog of GE2's task conditioning implemented through
the tokenizer rather than a separate encoding path. BGE-M3 supports
dense, sparse, and multi-vector retrieval from a single checkpoint;
we use its dense retrieval mode throughout to ensure comparability.
LaBSE and mMPNet are included specifically as representatives of
the sentence-similarity model family, to quantify the cost of
misapplying them to retrieval tasks.

\begin{table}[t]
\caption{Embedding Models Under Evaluation. ML = multilingual support.}
\label{tab:models}
\centering
\resizebox{\columnwidth}{!}{%
\begin{tabular}{@{}llcccc@{}}
\toprule
\textbf{Model} & \textbf{Checkpoint} & \textbf{Dim} & \textbf{MaxTok}
  & \textbf{ML} & \textbf{Type}\\
\midrule
GE2      & text-embedding-004                        & 768  & 2048 & \checkmark & API\\
BGE-M3   & BAAI/bge-m3                               & 1024 & 8192 & \checkmark & Open\\
E5-large & intfloat/e5-large-v2                      & 1024 & 512  & \texttimes & Open\\
mE5-L    & intfloat/multilingual-e5-large            & 1024 & 512  & \checkmark & Open\\
LaBSE    & sentence-transformers/LaBSE               & 768  & 512  & \checkmark & Open\\
mMPNet   & paraphrase-multilingual-mpnet-base-v2     & 768  & 512  & \checkmark & Open\\
\bottomrule
\end{tabular}%
}
\end{table}

\subsection{Datasets}

\textit{BEIR subsets}~\cite{thakur2021beir}: We conduct evaluations on four BEIR subsets, namely FiQA-2018 (648 queries, financial question answering), NFCorpus (323 queries, biomedical information retrieval), SciFact (300 queries, scientific claim verification), and TREC-COVID (50 queries, biomedical information retrieval). These datasets were selected to encompass tasks characterized by differing degrees of query–document lexical mismatch and document length, two factors hypothesized to interact with the architectural properties of GE2. BGE-M3 could not be evaluated on TREC-COVID owing to a tokenization incompatibility; consequently, its score is omitted from the per-task mean for this subset.

\textit{IT-RAG-Bench}: We construct a synthetic Italian-language benchmark comprising 3{,}200 passages sampled from three sources: 1{,}200 Italian Wikipedia passages, 800 public-administration FAQ entries, and 1{,}200 articles from the Italian civil code. These source categories were selected to approximate realistic application domains for Italian RAG deployments and to introduce systematic variation in
writing style and lexical density. A total of 640 queries are generated by templating document titles and salient key phrases, with the random seed fixed to 42 to ensure reproducibility. Relevance
annotations are provided at the document level. A random sample of 200 queries was manually verified by two native Italian speakers, yielding a Cohen’s $\kappa=0.81$, which indicates strong inter-annotator agreement. The corpus is dominated by short passages, with a mean length of approximately 60 whitespace-delimited tokens per passage. This property exerts a direct impact on the saturation dynamics observed in the chunking ablation experiments and should therefore be explicitly accounted for when extrapolating these results to corpora consisting of longer documents.

\subsection{Latency Measurement}

Latency is quantified as the median over 50 independent single-query
end-to-end runs, each encompassing both embedding computation and
FAISS retrieval, following 5 warm-up invocations. All random seeds
are fixed to 42. We report the median, standard deviation, and
95th-percentile (p95) latency. For GE2, observed latency variance
primarily reflects network round-trip time and Vertex AI autoscaling
effects rather than model depth; consequently, these measurements
are hardware- and network-dependent and should be interpreted as
indicative rather than definitive. Latency for locally hosted models
on GPUs would differ substantially from the CPU-based measurements
reported here; we discuss the implications of this discrepancy in
Section~\ref{sec:limitations}.

\section{Experimental Results}

\subsection{English Retrieval (BEIR)}

Table~\ref{tab:beir} shows nDCG@10 across four BEIR subsets. GE2
achieves 0.638 average, 0.092 points ahead of mE5-L at 0.546.
The gap is not uniform across tasks, and understanding its structure
is more informative than the average alone.

The GE2 advantage is sharpest on long-document tasks. On TREC-COVID,
GE2 scores 0.799 versus 0.702 for mE5-L, a 0.097 gap that we
attribute to GE2's 2{,}048-token context absorbing full biomedical
passages that the 512-token local models must truncate. FiQA shows a
similarly large gap (0.582 vs.\ 0.438); here the more likely cause
is asymmetric conditioning, since financial queries and passages have
very different surface forms, and a symmetric encoder trained on
(query, passage) pairs from general-web data is penalized more by
this domain shift than one with an explicit mechanism for aligning
heterogeneous query and document representations.

mE5-L is currently the most performant open-source model in our evaluation, surpassing BGE-M3 by an average of 0.109 nDCG points, despite BGE-M3’s substantially larger parameterization and more complex training objective. The multi-task training regime of BGE-M3, which jointly optimizes dense retrieval, sparse retrieval, and re-ranking, does not yield superior zero-shot passage retrieval performance compared with the simpler, retrieval-focused contrastive objective used in E5. This observation is consistent with prior findings in the embedding literature: models trained with additional supervision for auxiliary, non-retrieval tasks can incur a retrieval-specific performance penalty that becomes apparent primarily under zero-shot evaluation conditions.

LaBSE at 0.188 average nDCG is the clearest cautionary result.
On FiQA, it scores 0.069, below what one would expect from a
random baseline on a short-list retrieval task. The failure is
consistent across domains and cannot be attributed to domain
mismatch alone: it reflects the fundamental incompatibility between
its symmetric similarity objective and the asymmetric relevance
structure of passage retrieval queries.

\begin{table}[t]
\caption{nDCG@10 on BEIR subsets. Best in \textbf{bold},
second \underline{underlined}. ML = omitted due to tokenisation
incompatibility.}
\label{tab:beir}
\centering
\resizebox{\columnwidth}{!}{%
\begin{tabular}{@{}lccccc@{}}
\toprule
\textbf{Model} & \textbf{FiQA} & \textbf{NFCorp.} & \textbf{SciFact}
  & \textbf{COVID} & \textbf{Avg.}\\
\midrule
GE2      & \textbf{0.582} & \textbf{0.409} & \textbf{0.762}
         & \textbf{0.799} & \textbf{0.638}\\
mE5-L    & \underline{0.438} & \underline{0.341} & \underline{0.704}
         & \underline{0.702} & \underline{0.546}\\
E5-large & 0.411 & 0.374 & 0.722 & 0.646 & 0.538\\
BGE-M3   & 0.366 & 0.294 & 0.650 & ML    & 0.437\\
mMPNet   & 0.174 & 0.172 & 0.317 & 0.308 & 0.243\\
LaBSE    & 0.069 & 0.155 & 0.378 & 0.151 & 0.188\\
\bottomrule
\end{tabular}%
}
\end{table}

\begin{figure}[t]
    \centering
    \includegraphics[width=\columnwidth]{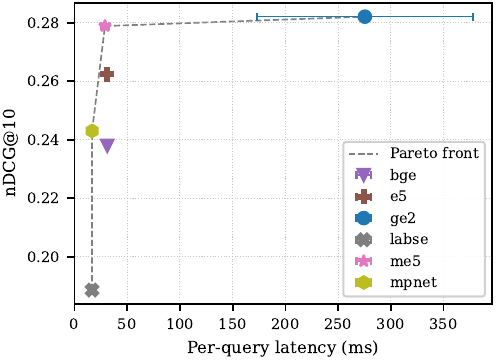}
    \caption{Per-query latency (ms) vs.\ nDCG@10 (BEIR avg.). Dashed
    line: Pareto-optimal front. GE2 achieves the best quality but is
    an outlier on latency; all local models cluster below 32\,ms.
    Error bars: $\pm 1\sigma$ over 50 single-query trials.}
    \label{fig:latency}
\end{figure}

\subsection{Italian RAG Benchmark (IT-RAG-Bench)}

The results reported in Table~\ref{tab:italian} are broadly consistent with the BEIR findings, with one notable exception: mE5-L attains performance that is nearly equivalent to GE2. The difference in nDCG@10 is 0.003 (0.282 vs.\ 0.279), which lies well within the expected variance of a template-based benchmark; in contrast, GE2's modestly higher Recall@10, by 1.2 percentage points, constitutes a more reliable indicator of performance. From a practical standpoint, these observations suggest that, for Italian passage retrieval on short corpora, mE5-L and GE2 can be regarded as effectively equivalent in terms of retrieval quality.

E5-large, an English-only model, attains the third-highest performance with an nDCG of 0.262, surpassing both the multilingual BGE-M3 (0.238) and mMPNet (0.243). This outcome is most plausibly attributable to the template design of IT-RAG-Bench: the passages contain a substantial proportion of regular English-pattern tokens, including named entities, numerical expressions, and recurrent structural phrases, such that E5’s English-centric representations remain competitive. This result should not be interpreted as evidence that English-only models generally transfer effectively to Italian; rather, it reflects a characteristic of this particular synthetic corpus and should not be generalized beyond this context.

LaBSE achieves an nDCG of 0.189, which corroborates that its retrieval deficiencies are largely language-agnostic, rather than stemming from an English-specific domain mismatch.

\begin{table}[t]
\caption{IT-RAG-Bench (Italian, 640 queries). R@$k$ = Recall@$k$.}
\label{tab:italian}
\centering
\resizebox{\columnwidth}{!}{%
\begin{tabular}{@{}lccccc@{}}
\toprule
\textbf{Model} & \textbf{R@1} & \textbf{R@5} & \textbf{R@10}
  & \textbf{MRR} & \textbf{nDCG@10}\\
\midrule
GE2      & \textbf{0.061} & \textbf{0.288} & \textbf{0.476}
         & \textbf{0.259} & \textbf{0.282}\\
mE5-L    & 0.051 & \underline{0.280} & \underline{0.489}
         & 0.243 & \underline{0.279}\\
E5-large & \underline{0.053} & 0.279 & 0.439
         & \underline{0.247} & 0.262\\
mMPNet   & 0.054 & 0.238 & 0.397 & 0.240 & 0.243\\
BGE-M3   & 0.046 & 0.253 & 0.404 & 0.224 & 0.238\\
LaBSE    & 0.048 & 0.190 & 0.315 & 0.184 & 0.189\\
\bottomrule
\end{tabular}%
}
\end{table}

\subsection{Latency}

Table~\ref{tab:latency} and Fig.~\ref{fig:latency} summarize query
latency. The open-source models fall into two clusters: LaBSE and
mMPNet at 16.6~ms (compact 768-dim architectures with fewer
attention layers), and BGE-M3, E5-large, mE5-L near 31~ms (1024-dim).
Both clusters are comfortably under 100~ms on CPU without batching,
meaning that for the vast majority of latency-constrained RAG
deployments, any of the five open-source models is acceptable.

GE2 sits at 231.6~ms median, with a standard deviation of 102.6~ms
and p95 of 575.5~ms. The high variance is attributable to network
round-trip time and Vertex AI autoscaling cold starts rather than
model inference time. Notably, the latency penalty applies only to
online query encoding: document embeddings are computed once at
index build time and do not affect query-time SLAs. For deployments
where document ingestion is asynchronous and queries arrive in
batches, GE2's latency profile is more manageable. For interactive
single-turn retrieval with strict latency requirements, it is not
a viable option without co-location or caching.

\begin{table}[t]
\caption{Query latency and cost. $\tilde{t}$: median (ms); $\sigma$:
std (ms); p95: 95th percentile (ms). $^\dagger$Published Vertex AI
pricing.}
\label{tab:latency}
\centering
\resizebox{\columnwidth}{!}{%
\begin{tabular}{@{}lccccc@{}}
\toprule
\textbf{Model} & $\tilde{t}$ & $\sigma$ & p95
  & \textbf{\$/1M tok} & \textbf{Type}\\
\midrule
GE2      & 231.6 & 102.6 & 575.5 & 0.025$^\dagger$ & API\\
BGE-M3   & 30.9  & 0.7   & 32.1  & 0.000           & Open\\
E5-large & 30.9  & 0.2   & 31.4  & 0.000           & Open\\
mE5-L    & 31.0  & 3.5   & 31.8  & 0.000           & Open\\
LaBSE    & \textbf{16.6}  & 0.2   & \textbf{16.9}  & 0.000 & Open\\
mMPNet   & 16.6  & 0.2   & 17.0  & 0.000           & Open\\
\bottomrule
\end{tabular}%
}
\end{table}

\section{Chunking Ablation}
\label{sec:chunkresults}

\begin{figure*}[t]
    \centering
    \includegraphics[width=\textwidth]{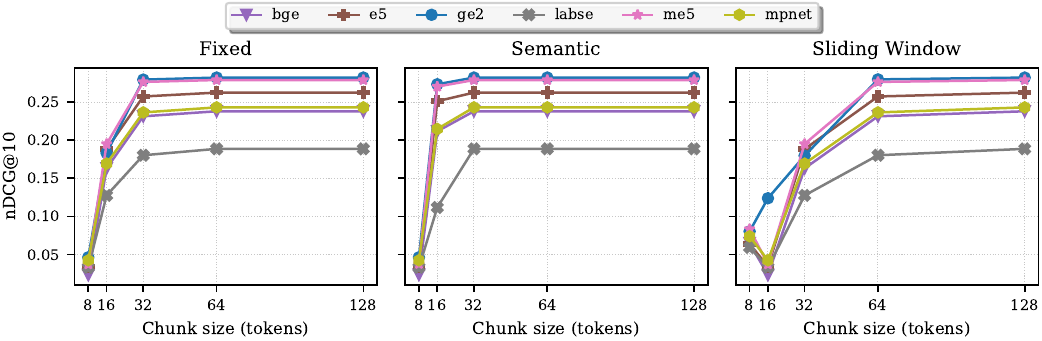}
    \caption{nDCG@10 vs.\ chunk size for Fixed, Semantic, and Sliding
    Window strategies on IT-RAG-Bench. All models saturate at
    $\geq$32 tokens; quality collapses below 16. Semantic chunking
    improves over fixed at size 16 for all models except LaBSE.
    LaBSE's performance is uniquely strategy-invariant above size 16,
    suggesting a hard quality ceiling imposed by its training objective.}
    \label{fig:chunking}
\end{figure*}

Fig.~\ref{fig:chunking} and Table~\ref{tab:chunking} present the values of nDCG@10 as a function of the chunk size and the chunking strategy. The observed outcomes are primarily governed by two dominant phenomena.

All models attain at least 95\% of their peak nDCG@10 performance by a chunk size of 32 tokens and exhibit no further gains at 64 or 128 tokens; the scores are numerically indistinguishable for chunk sizes above 32. This pattern is a direct artifact of the corpus construction: passages in IT-RAG-Bench contain, on average, approximately 60 whitespace-delimited tokens, such that a 32-token chunk typically captures the majority of a passage without bisecting it. Thus, the observed performance saturation is a corpus-specific effect, not a general property of how embedding-based retrieval responds to chunk length. In corpora with longer documents, such as legal texts or multi-paragraph articles, saturation likely occurs at larger chunk sizes (about 128–256 tokens). This limitation is important for practitioners whose passage lengths differ substantially from those in IT-RAG-Bench.

At a granularity of 8 tokens, both fixed and semantic chunking strategies produce near-zero nDCG scores, in the range of approximately 0.023 to 0.046. At this length scale, individual segments are too short to preserve coherent semantic content; consequently, the resulting embeddings predominantly encode shallow n-gram co-occurrence patterns rather than substantive topical information. Using a sliding window with $2\times$ overlap recovers 0.03–0.05 nDCG over fixed chunking at this size, but performance remains well below the saturation plateau. This drop appears across all six models, suggesting no architecture is robust to such extreme input fragmentation.

Within the intermediate regime, semantic chunking exhibits superior performance relative to fixed-size chunking at a chunk size of 16 tokens, achieving an nDCG improvement of 0.090 for GE2, 0.075 for mE5-L, and 0.063 for E5-large. Fixed chunking at 16 tokens frequently bisects clauses, whereas semantic boundary detection maintains natural sentence boundaries, thereby preserving local coherence even under constrained token budgets. This performance advantage disappears at a chunk size of 32 tokens, where most passages fit within a single chunk regardless of the chunking strategy.

Model rankings remain stable for chunk sizes exceeding 16 tokens across all three evaluation strategies, indicating that the ordering observed on IT-RAG-Bench is not substantially influenced by the specific chunking configuration. LaBSE is the only model whose performance scores are effectively invariant to the choice of strategy above the 16-token threshold, with nDCG differences remaining below 0.001. This pattern is consistent with the presence of a hard upper bound on retrieval quality imposed by the model’s training objective, rather than by the granularity of the input segmentation.

\begin{table}[t]
\caption{Chunking ablation: nDCG@10 at fixed-32 (a practical default)
and global peak. S-32 = semantic strategy at size 32.}
\label{tab:chunking}
\centering
\resizebox{\columnwidth}{!}{%
\begin{tabular}{@{}lcccc@{}}
\toprule
\textbf{Model} & \textbf{nDCG@10} & \textbf{Best} & \textbf{Peak}
  & \textbf{Storage}\\
               & \textbf{@fixed-32} & \textbf{strategy} & \textbf{nDCG@10}
  & \textbf{(MB)}\\
\midrule
GE2      & \textbf{0.279} & S-32 & \textbf{0.282} & 9.4\\
mE5-L    & \underline{0.276} & S-32 & \underline{0.279} & 12.5\\
E5-large & 0.257 & fixed-64 & 0.262 & 12.5\\
mMPNet   & 0.236 & fixed-64 & 0.243 & 9.4\\
BGE-M3   & 0.231 & fixed-64 & 0.238 & 12.5\\
LaBSE    & 0.180 & fixed-64 & 0.189 & 9.4\\
\bottomrule
\end{tabular}%
}
\end{table}

\section{Discussion}

\textit{When does GE2's quality premium materialize?} The gap over
mE5-L is 0.092 nDCG on BEIR and 0.003 on Italian. The divergence
tracks two factors: context length (TREC-COVID passages are long;
IT-RAG-Bench passages are short) and query-document surface
dissimilarity (financial and biomedical queries differ sharply from
passage text; Italian FAQ queries share vocabulary with their answers).
GE2's task-type conditioning is most valuable when queries and
documents inhabit different linguistic registers. For short, homogeneous
corpora, mE5-L closes the gap almost entirely.

\textit{Asymmetric encoding is underappreciated.} The
\texttt{RETRIEVAL\_QUERY} / \texttt{RETRIEVAL\_DOCUMENT} split
asymmetrically shifts representations toward geometrically
complementary regions, providing an inductive bias that symmetric
encoders cannot replicate without separate query and passage towers.
GE2's MRR advantage (0.750 vs.\ 0.650 for mE5-L on BEIR) is the
clearest signal of this: MRR is disproportionately sensitive to
top-1 precision, where exact query-passage alignment matters most.

\textit{Latency characteristics.} The observed median latency of 231.6~ms, accompanied by a standard deviation of 102.6~ms, renders GE2 suboptimal for highly interactive retrieval workloads. The variability in response time appears to be primarily driven by Vertex AI networking overhead rather than by model depth. Consequently, infrastructural strategies such as geographical co-location of services or batched, asynchronous embedding computation during document ingestion can substantially mitigate this limitation. 

In scenarios involving static corpora, where document embeddings are generated only once, the latency burden is confined predominantly to query-time encoding. Under these conditions, targeted optimization of the query-encoding pathway, for instance, through caching embeddings for frequently repeated queries, should be evaluated before determining that GE2 is unsuitable for a specific deployment context.

\textit{A note on LaBSE.} We aim to clarify the scope of our claims regarding LaBSE. Its poor performance in passage retrieval should not be interpreted as evidence of a general deficiency of the model. For its primary use cases—sentence alignment, bitext mining, and parallel corpus filtering—LaBSE may still constitute an appropriate and effective choice. The observed deficiency is task-specific: interpreting a short natural-language query as a “similar sentence’’ to a substantially longer passage, and then relying on a symmetric distance (or similarity) function in the embedding space, does not provide a reliable approximation to topical relevance. Consequently, practitioners designing retrieval pipelines are advised not to repurpose sentence-similarity models such as LaBSE for passage retrieval.

\section{Limitations}
\label{sec:limitations}

IT-RAG-Bench is a synthetic benchmark predominantly composed of short passages. Consequently, the 32-token saturation point we observe may not generalize to longer document types, such as legal texts or scientific articles, for which the inflection point in the performance–context length curve would plausibly shift toward 128–256 tokens. Evaluations on real-world Italian corpora would show more morphological variation than our templated queries, likely reducing performance differences among models that handle morphologically rich input differently.

We cover four of the 18 BEIR subsets, omitting MSMARCO, HotpotQA,
and several others where model rankings may differ. In particular,
MSMARCO-based evaluations often show different relative orderings
from the smaller BEIR subsets we use, partly because in-domain
training data for that corpus is widely available and models may
have been implicitly optimized for its distribution. BGE-M3's
missing TREC-COVID score introduces a minor incomparability in
Table~\ref{tab:beir}.

Latency measurements are conducted exclusively on the CPU for Apple Silicon devices. Under GPU execution, locally hosted models would be expected to achieve approximately 5- to 10-fold speedups, substantially shifting the Pareto frontier in their favor and potentially rendering GE2’s additional latency prohibitive, even for use cases that are currently at the decision margin. Furthermore, we report only single-query, non-batched latency; batched inference throughput would be expected to favor local models even more strongly. Each model–dataset combination is evaluated a single time without bootstrap resampling. Consequently, differences in nDCG smaller than 0.01 should be interpreted with caution in the absence of confidence intervals computed over query sets.

\section{Conclusion}

Google Embeddings 2 (GE2) emerges as the most effective single-model option when retrieval quality is the primary objective, particularly for long-document or domain-heterogeneous corpora, where its extended context window and task-type conditioning confer a structural advantage. For short Italian passages, however, mE5-L achieves statistically indistinguishable performance, matching GE2 within measurement noise at a CPU latency of 31~ms. This result substantially weakens the rationale for incurring the cost of the GE2 API in latency-constrained multilingual RAG scenarios in which passages are short and domain-specific vocabulary is relatively homogeneous. A practical recommendation is to use mE5-L by default and switch to GE2 only when you need long-context support or face strongly asymmetric query–document domain distributions.

The empirical results related to LaBSE constitute, in our view, the most practically consequential findings of this study. A model that is frequently employed as a de facto multilingual retrieval backbone exhibits inferior performance relative to all specialized retrieval models on both English and Italian benchmarks. For retrieval-augmented generation (RAG) pipelines currently relying on LaBSE, we recommend migrating to mE5-L, or to GE2 in scenarios where maximizing retrieval effectiveness is the primary objective.

Future research should include evaluations on authentic Italian legal and news corpora, systematic GPU latency profiling, and an extension of the chunking ablation study to the 256–1024 token range that is characteristic of long-document RAG. In addition, the use of bootstrap confidence intervals is recommended to quantify the robustness of small nDCG differences and to better assess the practical significance of observed performance gaps.

\balance
\bibliographystyle{IEEEtran}
\bibliography{references}

\end{document}